%% file: main.tex
\documentclass[lettersize,journal,article]{IEEEtran}
\usepackage{amsmath,amsfonts}

\usepackage{algpseudocode}
\usepackage{lipsum}
\usepackage{dsfont}
\usepackage{array}
\usepackage[caption=false,font=normalsize,labelfont=sf,textfont=sf]{subfig}
\usepackage{textcomp}
\usepackage{stfloats}
\usepackage{multirow}
\usepackage[flushleft]{threeparttable}
\usepackage{url}

\usepackage{verbatim}
\usepackage{graphicx}
\usepackage{cite}
\usepackage{amsmath}
\usepackage{algpseudocode}
\usepackage{algorithm}
\usepackage{algorithmicx}
\usepackage{todonotes}

\begin{document}
\bstctlcite{IEEEexample:BSTcontrol}

\title{CHASE: A Causal Hypergraph based Framework for Root Cause Analysis in Multimodal Microservice Systems}

\author{
    \input{author.tex}

}

\markboth{Journal of \LaTeX\ Class Files,~Vol.~14, No.~8, April~2024}%
{Shell \MakeLowercase{\textit{et al.}}: A Sample Article Using IEEEtran.cls for IEEE Journals}


\maketitle

\begin{abstract}
In recent years, the widespread adoption of distributed microservice architectures across the industry has greatly increased the demand for improved system availability and robustness. Enterprise-level microservice systems often involve complex service invocation paths and system dependencies, making it challenging to quickly identify anomalies during service interactions, which inevitably hinders normal system operations and maintenance. In this paper, we propose CHASE—a \underline{\textbf{C}}ausal \underline{\textbf{H}}ypergraph gr\underline{\textbf{A}}ph ba\underline{\textbf{S}}ed fram\underline{\textbf{E}}work—for root cause analysis in microservice systems. CHASE is designed to handle multimodal data sources, including traces, logs, and system monitoring metrics. Specifically, CHASE begins by encoding each type of input data—such as traces, logs, and metrics—into representative embeddings. These embeddings are then integrated into a unified multimodal invocation graph that captures the relationships among services. Anomaly detection is subsequently applied to each service instance node through attentive heterogeneous message passing, leveraging contextual information from adjacent metric and log nodes. Finally, CHASE constructs a causal hypergraph, where hyperedges represent inferred causal relationships among anomalies, and utilizes this structure to accurately localize root causes within the microservice system.
We evaluate the proposed framework on two public microservice datasets with distinct attributes and compare with seven benchmark methods. The results show that CHASE achieves the average performance gain up to 36.2\%(A@1) and 29.4\%(Percentage@1), respectively to its best counterpart.
\end{abstract}


\input{introduction.tex}

\input{relatedwork}

\input{methodology}

\input{experiment}

\section{Conclusion}\label{section5}
For complex microservice systems, conducting a root cause analysis is essential in identifying and addressing the underlying issues that hinder the optimal performance of the system.  
To tackle this challenge, we propose a causal hypergraph based framework named CHASE to accurately locate the root cause instances of failures, facilitating effective troubleshooting and system maintenance. Starting with modeling microservice systems with graph representations, we apply encoders and embedding layers combined with positional encoding to encode the multimodal data of trace, log and metrics into an invocation graph. 
CHASE employs heterogeneous message passing on invocation graphs to tackle instance-level anomaly detection. Next, it constructs a hypergraph with each hyperedge capturing the causality flow for each instance. Finally, the designated hypergraph module locates the root cause anomaly with a detection head applied to each updated node embedding after hypergraph convolution. We evaluate CHASE using two real-world datasets with seven baseline methods, and the evaluation results confirm the superiority of our proposed framework. For the future work,  hypergraph learning techniques on modelling spatiotemporal graphs can be applied to capture the causality structure of the trace, which is dynamically evolving since the whole trace typology is gradually formed by sequential invocations ranging in a certain time interval. Additionally, a single powerful autoregressive encoder that is intrinsically trained on multimodal data, such as the large language model, can substitute the dedicated log and metric encoders of CHASE, in order to enhance the quality of embeddings, as well as robustly unify the information of metrics and log.


\bibliographystyle{IEEEtran}
\bibliography{reference}

\appendices




\end{document}

%% file: author.tex
Ziming Zhao\(^*\),
Zhenwei Wang\(^*\),
Tiehua Zhang\textsuperscript{\dag}, \IEEEmembership{Member, IEEE},
Zhishu Shen, \IEEEmembership{Member, IEEE},
Hai Dong, \IEEEmembership{Senior Member, IEEE},
Zhen Lei,
Xingjun Ma, \IEEEmembership{Member, IEEE},
Gaowei Xu, \IEEEmembership{Member, IEEE},
Zhijun Ding, \IEEEmembership{Senior Member, IEEE}, 
Yun Yang, \IEEEmembership{Senior Member, IEEE} 

\thanks{ \textsuperscript{\dag}\textit{Corresponding author: Tiehua Zhang}}
\thanks{ \textsuperscript{*}\textit{Ziming Zhao and Zhenwei Wang are equally contributed}}
\thanks{Ziming Zhao is with the NEC Laboratories America, United States (e-mail: zhziming1@gmail.com).}

\thanks{Zhenwei Wang is with the School of Computer Science, University of Nottingham Nningbo China, Ningbo, China (e-mail: \ zhenwei.wang@nottingham.edu.cn).}

\thanks{Tiehua Zhang and Zhijun Ding is with the School of Computer Science and Technology, Tongji University, Shanghai, China (e-mail: \{tiehuaz, dingzj\}@tongji.edu.cn ).}

\thanks{Zhishu Shen and Zhen Lei is with the School of Computer Science and Artificial Intelligence, Wuhan University of Technology, Wuhan, China (e-mail: z\_shen@ieee.org, studentlz@whut.edu.cn).}

\thanks{Hai Dong is with the School of Computing Technologies, RMIT University, Melbourne, Australia (e-mail: hai.dong@rmit.edu.au).}

\thanks{Xingjun Ma is with the School of Computer Science, Fudan University, Shanghai, China (e-mail: xingjunma@fudan.edu.cn).}

\thanks{Gaowei Xu is with the College of Electronic and Information Engineering
Tongji University, Shanghai, China (e-mail: xugaowei@tongji.edu.cn).}

\thanks{Yun Yang is with the School of Science, Computing and Engineering Technologies, Swinburne University of Technology, Melbourne, Australia (e-mail: yyang@swin.edu.au).}

%% file: introduction.tex
\section{Introduction}
\IEEEPARstart{M}{icroservice} architecture has recently surged in popularity due to its significant advantages such as enhancing flexibility,
scalability, and fault tolerance for contemporary industrial service-oriented systems~\cite{li2021practical, baboi2019dynamic, ramu2023performance}. This architectural design involves a collection of small, independent, and loosely coupled application instances, where each application delivers a specific business function~\cite{zhang2019esda}. Specifically, each microservice is independently developed, deployed, and managed by a dedicated team, with communication facilitated through APIs and protocols such as REST, HTTP, and messaging systems. 
However, fault localization and analysis in microservice systems often pose significant challenges 
as microservices are interconnected and each method or function call might trigger multiple distributed service invocations, either synchronously or asynchronously~\cite{peng2022large,liu2021microhecl,chen2014causeinfer}. Operations engineers are often overwhelmed by the need to manually collect data from diverse sources during system failures. This process involves managing intricately structured function call information, semi-structured log text, and system performance metrics, thereby complicating the identification and resolution of issues. As the complexity of systems continues to grow, there is an increasing emphasis on integrating Root Cause Analysis (RCA) methods to timely uncover the underlying reasons behind issues or failures~\cite{ikram2022root,brandon2020graph,kim2013root}.

\begin{figure*}[t!]
\includegraphics[width=0.98\linewidth]{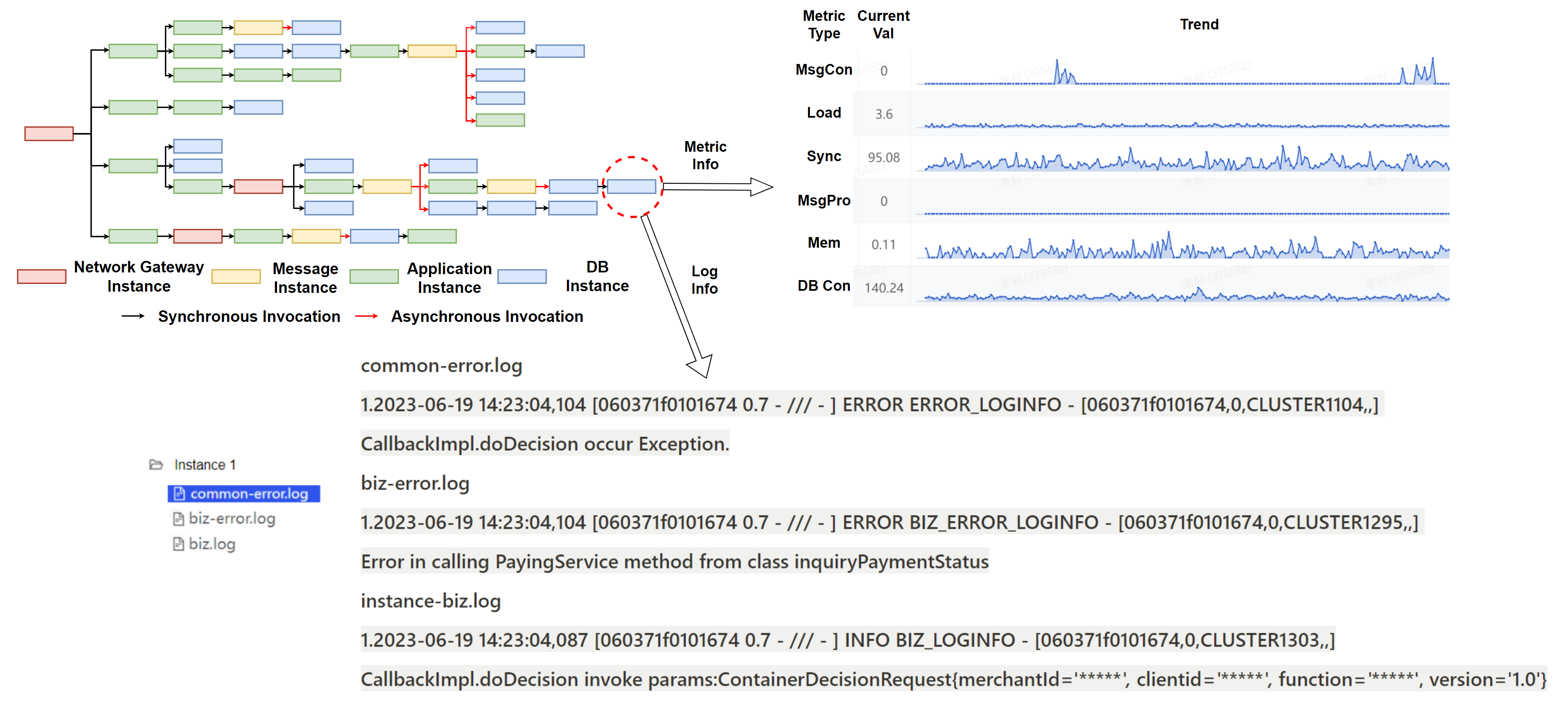}
\centering
\caption{An exemplary service invocations in the microservice system with multimodal data, including trace, monitoring metrics and log data} 
\label{1-1}
\end{figure*}
 
Traditional RCA methods only offer suggestive clues, necessitating detailed manual analysis to determine the exact root cause~\cite{soldani2022anomaly}. To date, numerous models have been proposed for root cause analysis. One class of approaches explores the feasibility of building causality graphs using merely trace information, which consists of service invocation chains from various service instances.
Based on such intuition, causal analysis algorithms such as PC~\cite{sole2017survey} and GES~\cite{maxwell2002optimal} are employed to infer the causal relationships between service instances in the graph. 
Following that, a line of research~\cite{lin2018microscope,sharma2013cloudpd,wang2018cloudranger,wu2020microrca,xin2023causalrca, zhang2023robust,rong2022locating,bhandari2018fault,liu2021microhecl,zhang2024trace,yu2021microrank} starts to integrate root cause analysis with machine learning models and graph neural networks (GNNs), from which the graph learning techniques are leveraged to model the pairwise bivariate correlation of instances connected with directed invocation edges. In these work, multimodal data of service instances are used as supplementary features to enrich features such as node embeddings and edge weights, while the uncovering of bivariate correlation between pairs of instance nodes are treated as the homogeneous graph learning task in the trace typology, leading to the following challenges:
\begin{enumerate}
\item \textbf{Joint processing of multimodal information and invoke topology.} Existing non-graph neural network frameworks primarily extract causal relationships from multimodal data to construct causal structure graphs but fail to fully utilize call topology information. While GNN frameworks adopt graph-based learning methods, they fall short in simultaneously capturing multimodal information, instance invoking topology, and causal propagation.

\item \textbf{Multivariate causal correlation capture among instances.} 
In microservice systems, each instance node can influence multiple downstream calls, and causal information may propagate across nodes several hops away. Traditional graph neural networks primarily focus on modeling pairwise local correlations along directed edges, making it challenging to capture such complex causal flows. 
\end{enumerate}

To tackle these issues, we propose a novel causal hypergraph-based framework named CHASE, which is applicable to inductively learn the root cause anomaly of microservice systems with the presence of trace, log, and metric data all together. Based on multimodal data, a heterogeneous invocation graph is first constructed using trace topology, with additional metric and log nodes connected to the source instance nodes where the information is gathered. Representative node embeddings are then generated using designated encoders for each multimodal data type. The procedure for instance-level anomaly detection is carried out using heterogeneous attentive message passing. By doing so, CHASE can accurately locate root cause instances with excellent performance by learning from the constructed hypergraph, with hyperedges representing causality propagation in the trace. We evaluate CHASE on two datasets with different attributes—one with static trace topology and one with dynamic trace topology. Our experimental results show that CHASE significantly improves fault localization accuracy compared to several state-of-the-art methods.


The main contributions of this paper are as follows:
\begin{enumerate}
\item We introduce CHASE, an innovative graph learning framework tailored to process multimodal data, including logs, metrics, and traces, specifically for root cause analysis. This framework integrates the encoding of multimodal information with the development of an invocation graph-based typology. We accomplish multimodal feature fusion by solving the instance-level anomaly detection with heterogeneous message passing. Hereby, we attain the modeling of multimodal data and information heterogeneity of microservice systems.
\item We achieve the root cause analysis task with hypergraph learning. CHASE captures the causality flow of the anomalies in microservice systems by constructing hypergraphs on the basis of typology, from which each hyperedge represents causality propagation along the invocation path. The multivariate causality corrleation between a set of instances is modelled with hypergraph convolution.
\item We conduct extensive experiments using two public datasets from microservice systems and compare the performance with a number of traditional and GNN-based baselines, demonstrating that our proposal can outperform comparative methods in the root cause analysis task. Apart from that, we release the source code\footnote{https://drive.google.com/file/d/11erha3k8FeA67z-sfKGReqHz66PpO6o4/view?usp=sharing} to advance the developement of this field.

\end{enumerate}

The rest of the paper is organized as follows: Section~\ref{section2}
introduces the related work including non-GNN based and GNN-based root cause analysis. Section~\ref{section3} explains the overall framework of CHASE and the implementation details including multimodal invocation graph construction, instance level anomaly detection with heterogeneous message passing and hypergraph causality learning for root cause analysis. In Section~\ref{section4}, we evaluate the performance of CHASE and extensively compare it with the state-of-the-art baselines on two distinct datasets. Section~\ref{section5} concludes the paper.

%% file: relatedwork.tex
\section{Related Work}\label{section2}
\subsection{Non-GNN based Root Cause Analysis}
The complexity of the microservice system has intensified the need for effective root cause analysis techniques to diagnose and resolve issues. 
In recent years, a line of research has focused on developing RCA methods for microservices systems~\cite{gholami2021comparative}. 
This section overviews some of the key research studies in this area.

Existing work mainly uses three categories of data sources: log-based~\cite{zhang2021cloudrca}, trace-based~\cite{li2021practical}, and metric-based~\cite{ma2020self}. 
Log-based RCA is a technique that analyzes service logs from different instances in the microservice system to identify potential root causes of issues, which naturally rely on the accurate text parsing techniques in the log and are often hard to work in real time.  
For instance, Zhang et al.~\cite{zhang2021cloudrca} propose a method for localizing operational faults that involve two steps: it first preprocesses system logs to generate high-quality features, and then uses machine learning model on these features to identify the root cause of operational faults. 
LogFlash~\cite{jia2021logflash} also integrates anomaly detection on logs as the main part of root cause analysis based on the calculation of deviation from normal log status. 
A common issue for log-based RCA is that these works often require offline efforts to extract key information in the log, and the performance of the log-based RCA is also largely dependent on the overall quality of the system logs.
Trace-based research utilizes the information through the complete tracing of the execution paths and then identifies root causes that occur along the way.  
TraceRCA~\cite{li2021practical} uses a tracing tool to collect trace data among service invocations in the microservice system and employ the decision tree to detect the root cause. However, using trace data alone is insufficient as trace data only presents information at service invocation level~\cite{kim2013root}.  
Apart from the abovementioned two categories, the metirc-based RCA is now widely studied in the research community.  
Most metric-based RCA research~\cite{lin2018microscope} employs the monitoring data (e.g., CPU and memory usage, network latency etc.) gathered from different service instances to establish causal graphs and deduce the underlying root causes, including MicroRCA~\cite{wu2020microrca} and CloudRanger~\cite{wang2018cloudranger}, while the former correlate application performance symptoms with the root cause, the latter conducts second-order random walk on impact graph to identify the problematic services.  
However, the metric-based root cause analysis method does not consider other types of information from microservice systems. TrinityRCL harness telemetry data of application-level, service-level, host-level, combined with metric-level to construct the causal graph with heterogeneity\cite{gu2023trinityrcl}.

\subsection{GNN based Root Cause Analysis}
Graph Neural Networks have demonstrated prowess in capturing intricate relationships within graph-structured data, i.e. data from non-Euclidean spaces. GNNs can be explained as low-pass filters for graph Fourier transform from the spectral perspective, and can also be regarded as a message passing mechanism from neighbor nodes under the spatial perspective. By leveraging popular GNN architectures such as Graph Convolution Network (GCN)\cite{kipf2016semi}, GraphSAGE\cite{hamilton2017inductive}, and Graph Attention Network (GAT)\cite{velickovic2017graph}, tasks relating with graph structural learning have been solved in fields of social networks, biology, and recommendation systems, etc~\cite{zhang2023adaptive,liu2024exploiting}. Due to the intrinsic nature of microservice systems where individual services are pairwise correlated, it is intuitive to model the dependencies between microservices with a graph representation. Thus GNNs can be employed to learn the patterns of microservice systems and facilitate root cause analysis.

Recent advancement of GNNs has unleashed great potential in analyzing the root cause problems, especially for cases relying on graph structure data.  
Owing to the natural compatibility with generated causal graphs, researchers started to leverage GNN to learn the failure propagation patterns in the graph, thus giving a stronger generalization ability compared with random walk based RCA frameworks on causal graphs generated with Peter-Clark(PC) algoritm~\cite{zhang2023robust}. 
CausalRCA~\cite{xin2023causalrca} designs a gradient-based causal structure learning to capture linear and non-linear causal relations in monitoring metrics and outputs a weighted causal directed acyclic graph (DAG). Diagfusion~\cite{zhang2023robust} reports the state-of-the-art RCA results using GNN.  
It combines deployment data and traces to build a dependency graph and uses GNN to generate the embeddings for each service instance, which are then used to achieve two-fold failure diagnosis, i.e., root cause fault localization and failure type determination.  A hierarchical causal network framework named REASON\cite{wang2023hierarchical} is applied to model the fault propagation of the microservice trace with both within-network and across-network causal relationships
for root cause localization. Random walk on the learnt causal network is further applied to locate a system fault.
It is worth noting that CausalRCA uses the encoder-decoder structure for unsupervised learning to derive the weight of edges, which is considered insufficient to represent the significance between causes and effects in the causal graph.  
Diagfusion, on the other hand, applies graph convolutional network for representation learning on invocation graphs in an end-to-end fashion, requiring a certain amount of labelled data.
Moreover, Eadro\cite{lee2023eadro} employs GAT to learn representations of multimodal information status. It integrates anomaly detection and root cause localization into a unified end-to-end model, facilitating knowledge sharing and joint training.

In all, non-GNN based frameworks focus more on mining causal correlation from multimodal information, such as metrics, so as to build up causal structure graphs representing causality flow, while GNN-based frameworks apply graph learning methodologies and unravel the RCA task into a trainable optimizing problem by minimizing the traning loss.
However, none of the aforementioned non-GNN based frameworks or GNN-based frameworks can tackle RCA task by capturing multimodal information, invocation typology as well as causality propagation simultaneously in an end-to-end trainable manner. Our proposed framework, CHASE, focuses on exploiting logs and metrics to generate pertinent and comprehensive embeddings, applies message passing in graph to learn the non-Euclidean data of invocation traces, models causality flow with hyperedges and presents an overall training loss making RCA task soluble in an end-to-end way.

%% file: methodology.tex
\section{CHASE Framework}\label{section3}
\begin{figure*}[t!]
\includegraphics[width=\linewidth]{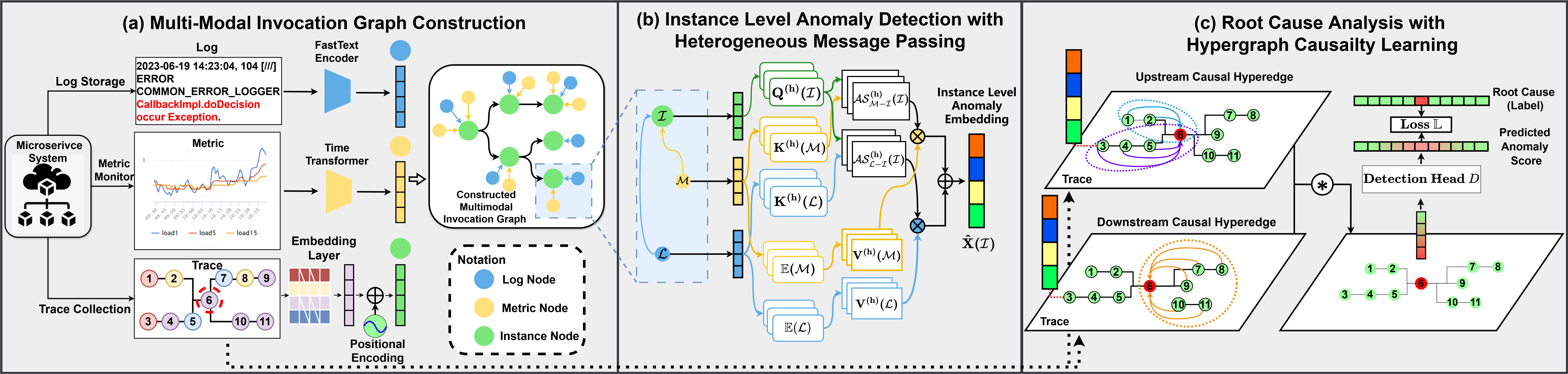}
\centering
\caption{Overall framework of CHASE}\label{3-0}
\label{pipeline}
\end{figure*}
In this section, we elaborate the CHASE framework with three stages, including invocation graph generation, instance anomaly detection with heterogeneous message passing, and root cause analysis with hypergraph learning. Fig.~\ref{3-0} depicts the overall architecture. Our proposed framework first constructs a graph model by integrating multimodal data and topological dependencies (Fig.~\ref{3-0}(a)). Then, a heterogeneous graph transformer (HGT) is employed to perform representation learning for instance nodes and their associated data nodes (Fig.~\ref{3-0}(b)), aiming to incorporate the heterogeneity attributes from multimodal data. Finally, hypergraph convolution is utilized to conduct root cause localization for the instance nodes (Fig.~\ref{3-0}(c)). We list all necessary notations used in this paper in Table~\ref{tab:notation} and provide a more detailed explanation of the three submodules in the following subsections.

\begin{table}[]
\centering
\caption{Key Notations}\label{tab:notation}
\begin{tabular}{c||l}
\hline Notation & Description \\
\hline \hline 
$\|$ & Concatenation \\
$\mid \cdot \mid$ & Set size \\
$\sigma(\cdot)$ & Activation function \\
$\mathcal{G}(\mathcal{I}, \mathcal{M}, \mathcal{L}, E)$ & Invocation graph modelling a microservice trace \\
$N(\cdot)$ & Neighbors of a node in $\mathcal{G}$ \\
$\mathcal{I}$ & Set of instance nodes \\
$\mathcal{M}$ & Set of metric nodes \\
$\mathcal{L}$ & Set of log nodes \\
$E$ & Set of edges \\
$\mathbb{E(\cdot)}$ & Encoded node embedding \\
$\mathds{1}_{\mathcal{I}}$ & One hot embedding for instance category \\
$H$ & Number of attention heads \\
$K^{(h)}(\cdot)$ & Key embedding of attention head with index $h$ \\
$Q^{(h)}(\cdot)$ & Query embedding of attention head with index $h$ \\
$V^{(h)}(\cdot)$ & Anomaly information of attention head with index $h$ \\
$\mathcal{AS}_{L-I}(\cdot)$ & Anomaly score between a pair of instance and log \\
$\mathcal{AS}_{M-I}(\cdot)$ & Anomaly score between a pair of instance and metric \\
$\Tilde{A}(\mathcal{I})^{(h)}$ & Nodewise normalized anomaly score of instance\\
$\mathcal{H}$ & Hypergraph incidence matrix\\
$D_{v}$ & Diagonal matrix of node degree\\
$D_{e}$ & Diagonal matrix of hyperedge degree\\
$D$ & Detection head for root cause analysis \\
\hline
\end{tabular}
\end{table}

\subsection{Multimodal Invocation Graph Construction}
To begin with, we formally define the concept of an invocation graph within the context of microservice systems, which is derived from a directed acyclic graph (DAG). A DAG is a directed graph that contains no cycles, meaning there is no path that starts at a given node and loops back to the same node after traversing multiple edges. An invocation graph is a special form of a DAG, where the edges represent the meta-relationships between instances, logs, and metrics.

In a microservice system, as illustrated in Fig.~\ref{1-1}, the handling of a request is typically represented by a microservice trace, which can be modeled as a directed acyclic graph (DAG). In this representation, instances serve as nodes, and each directed edge \(e(u, v)\) signifies that instance \(v\) is invoked by instance \(u\). There are multiple types of service instances, including application instances, message instances, gateway instances, database instances. The instances and invocation edges on a trace are extracted and assembled to obtain the complete microservice invocation graph to represent business processes. Generally, each instance continuously generates logs during its runtime, while the DevOps platform concurrently records metrics related to the instance's container and physical server.

This aforementioned multimodal trace information is modeled by the invocation graph denoted with $\mathcal{G}(\mathcal{I}, \mathcal{M}, \mathcal{L}, E)$, where $\mathcal{I}$ denotes the set of all instances in the trace, $\mathcal{M}$ denotes the set of monitored metrics of all instances, and $\mathcal{L}$ denotes all printed logs of the trace, including warnings and errors. We then apply directed edges $e(\mathcal{L}, u) \in E$ and $e(\mathcal{M}_n, u) \in E$ to represent the runtime records of metric and logs of instance node $u\in\mathcal{I}$. Notice that each instance is monitored with multiple types of metrics so that there may exist multiple metric-instance edges for a single instance, while we encode all the log information into one node per instance to reduce $|\mathcal{L}|$ for computational efficiency on the multimodal invocation graph.

Note that the raw feature of log nodes are natural language texts with semantic meaning, features of metric nodes are composed of real valued time series data and the instance nodes have categorical features. Consequently, an invocation graph that includes log, metric, and instance nodes can be regarded as a graph containing multimodal information. Hence, we apply respective encoders to encode the feature of log, metric and instance nodes into embeddings denoted as $\mathbb{E}(\mathcal{L}), \mathbb{E}(\mathcal{M}), \mathbb{E}(\mathcal{I})$, respectively:

\begin{gather}
\begin{split}
\mathbb{E}(\mathcal{L}) &= \textbf{FastTextEncoder}(\mathcal{L}) \\
\mathbb{E}(\mathcal{M}) &= \textbf{TimeSeriesEncoder}(\mathcal{M}^{T})[:, -1] \\
\mathbb{E}(\mathcal{I}) &= \mathds{1}_{\mathcal{I}}\Theta_e + P(k, d)
\end{split}
\end{gather}

For log nodes, we follow the work from \cite{zhang2023robust} which applies FastText~\cite{joulin2016bag} in encoding the extracted log templates into text embeddings. For metric nodes, we apply the trivial time series transformer provided by Huggingface\footnote{https://huggingface.co/blog/time-series-transformers}, and take the output embedding from the last timestamp as the encoded metric embedding. For instance nodes, the one-hot embedding $\mathds{1}_{\mathcal{I}}$, representing the category of the specific instance node, is projected with a learnable matrix $\Theta_e$. And we combine the information of the temporal invocation order of each instance in the trace by applying positional encoding. We denote positional encoding as $P(k, i)$ where $i$ denotes the index of a single feature in all feature dimensions $d$, $k$ denotes the temporal order in invocation and $n$ is a hyperparameter set to $20000$ in defualt, as follows:

\begin{gather}\label{eq:positional}
P(k, i)=\begin{cases}
    \sin{\frac{k}{n^{i/d}}}, & \text{if i $\text{mod}2 \equiv 0$}\\
    \cos{\frac{k}{n^{i/d}}}, & \text{if i $\text{mod}2 \equiv 1$}.
\end{cases}
\end{gather}

\subsection{Instance Level Anomaly Detection with Heterogeneous Message Passing}
After the whole encoding process where all the log node embeddings, metric node embeddings and instance node embeddings are encoded into $\mathbb{E}(\mathcal{L})$, $\mathbb{E}(\mathcal{M})$, $\mathbb{E}(\mathcal{I})$, we construct the multimodal invocation graph. The task of instance level anomaly detection is a conventional procedure in root cause analysis. In this section, we handle this task based on the intrinsic heterogeneity of the invocation graph inspired by the architecture design of heterogeneous graph transformer.

\subsubsection{Attentive Anomaly Score}
Starting with the subgraph composed of a single instance node and its related log and metric nodes, we model all the output logs with a single log node $\mathcal{L}$, while we model different types of monitored metrics with several metric nodes $n$, denoted as $\mathcal{M}_n$. We first project all the encoded embeddings into a specific vector space for anomaly detection. For log and metric nodes, we perform the following linear transformation. 
\begin{gather}
\begin{split}
K^{(h)}(\mathcal{L}) &= \mathbb{E}(\mathcal{L}) W^{(h)}_{\textbf{Log}} \\
K^{(h)}(\mathcal{M}_n) &= \mathbb{E}(\mathcal{M}_n) W^{(h)}_{\textbf{Metric}_n} \\
Q^{(h)}(\mathcal{I}) &= \mathbb{E}(\mathcal{I}) W^{(h)}_{\textbf{Instance}} 
\end{split}
\end{gather}
where $W^{(h)}_{\textbf{Log}}$, $W^{(h)}_{\textbf{Metric}_n}$, $W^{(h)}_{\textbf{Instance}}$ are learnable weight matrices of dimension $d\times d/H$, where $d$ is the inital embedding dimension from our multimodal encoders and $H$ is the total number of attention heads, $h$ denotes the index of a specific attention head. 

For each instance node in the microservice system, multiple types of runtime metrics and logs are continuously produced and monitored by the DevOps platform. It is crucial to identify which of these monitored records are most representative of the detected instance-level anomaly. We achieve this by quantifying the anomaly score, denoted as $\mathcal{AS}$, with the heterogeneous mutual attention weight between the instance node and all its neighbors in the invocation graph.
\begin{gather}
\begin{split}
\mathcal{AS}_{\mathcal{L}-\mathcal{I}}^{(h)}(\mathcal{I}) &= K^{(h)}(\mathcal{L})A_{\mathcal{L}-\mathcal{I}}Q^{(h)}(\mathcal{I})^T\times\frac{\phi_{\mathcal{L}}}{\sqrt{d}} \\
\mathcal{AS}_{\mathcal{M}_n-\mathcal{I}}^{(h)}(\mathcal{I}) &= K^{(h)}(\mathcal{M}_n)A_{\mathcal{M}_n-\mathcal{I}}Q^{(h)}(\mathcal{I})^T\times\frac{\phi_{\mathcal{M}_n}}{\sqrt{d}}
\end{split}
\end{gather}

Both $A_{\mathcal{L}-\mathcal{I}}$, $A_{\mathcal{M}_n-\mathcal{I}}$ are of dimension $d/H \times d/H$, which represents the projection matrix for calculating the attention weight between the instance node and its multimodal neighbors. This matrix is designed to be shared among all attention heads. The parameters \(\phi_\mathcal{L}\) and \(\phi_{\mathcal{M}_n}\) represent the prior significance assigned to different types of metrics and logs. These parameters are learnable and can be initially set to ones, reflecting our initial assumption that all types of logs and metrics are equally important for identifying instance-level anomalies.

The anomaly score is normalized for each instance node with all its neighbors.
\begin{gather}\label{attnweight} 
\begin{split}
\Tilde{A}^{(h)}(\mathcal{I}) = \underset{\forall \mathcal{L},\mathcal{M}_n \in N_\mathcal{I}}{\text{Softmax}}\{\mathcal{AS}_{\mathcal{L}-\mathcal{I}}^{(h)}(\mathcal{I})
\mathbin\Vert \mathcal{AS}_{\mathcal{M}_n-\mathcal{I}}^{(h)}(\mathcal{I})\}
\end{split}
\end{gather}
The instance anomaly score for attention head $h$ is of dimension $1\times |N_\mathcal{I}|$, where $N_\mathcal{I}$ denotes the set of all neighbors of node $I$ and $\sum \Tilde{A}^{(h)}(\mathcal{I}) = \mathbf{1}$ in accordance with the normalized attention weight.

\subsubsection{Anomaly Information Passing}
CHASE further on extracts the anomaly information from the metric nodes and log nodes by performing another linear transformation with their encoded embeddings.
\begin{gather}
\begin{split}
V^{(h)}(\mathcal{L}) &= \mathbb{E}(\mathcal{L}) W^{\prime(h)}_{\textbf{Log}}\\
V^{(h)}(\mathcal{M}_n) &= \mathbb{E}(\mathcal{M}_n) W^{\prime(h)}_{\textbf{Metric}_n}
\end{split}
\end{gather}
In order to extract the anomaly detection information from all the attention heads, we concat all the embeddings together and form:
\begin{gather}
V^{(h)}(\mathcal{I}) = V^{(h)}(\mathcal{L})\underset{\forall \mathcal{M}_n\in N_{\mathcal{I}}}{\mathbin\Vert} V^{(h)}(\mathcal{M}_n)
\end{gather}
where $V^{(h)}(\mathcal{I})$ represents the anomaly information of instance node $\mathcal{I}$ from all its neighbors $N_\mathcal{I}$ and with the dimension of $|N_\mathcal{I}| \times d/H$. Since each attention head will extract the anomaly detection information from a separate projection field, $V^{(h)}(\mathcal{I})$ will be stacked up to the size of $H \times |N_\mathcal{I}| \times d/H$, which will be weighted summed by $\Tilde{A}^{(h)}(\mathcal{I})$ so as to gather all the anomaly information from the neighbors of instance node $I$ based on the significance of the attentively learnt anomaly score.

\subsubsection{Instance Level Anomaly Embedding Update}
We aggregate all the anomaly information based on its attention weight calculated from Equation~\ref{attnweight} by:

\begin{gather}
\begin{split}
\hat{X}(\mathcal{I}) = \underset{h\in [1, H]}{\mathbin\Vert}&\{\Tilde{A}^{(h)}(\mathcal{I})V^{(h)}(\mathcal{I})\}
\end{split}
\end{gather}

$\hat{X}(\mathcal{I})$ denotes the instance level anomaly embedding learnt from the heterogeneous invocation graph. Each instance node will gather anomaly information from its monitored log and metric neighbor nodes, and the instance node embedding is updated with this learnt anomaly embedding, which reflects the procedure of anomaly detection.
\begin{gather}\label{anomaly_emb}
\Tilde{X} = (1-\gamma) \hat{W}_I\sigma(\hat{X}(\mathcal{I})) + \gamma \mathbb{E}(\mathcal{I})
\end{gather}

$\hat{W}_I$ is applied to project the anomaly information embedding back to the same vector space with the instance node features $\mathbb{E}(\mathcal{I})$ after being activated. $\gamma$ is the hyperparameter representing our prior knowledge on how significantly an instance-level anomaly will result in the root cause of a trace-level anomaly. The whole framework of instance level anomaly detection is delineated in Fig.~\ref{pipeline}(b).

\subsection{Hypergraph Causality Learning for RCA}
Lastly, CHASE perform root cause analysis with constructed hyperedges representing causality flow on the invocation graph, where heterogeneous anomaly information is propagated to each instance node.
Let $\mathcal{G}=(V, E)$ be a hypergraph with incidence matrix $\mathcal{H}$. $V$ denotes the vertices of the hypergraph, $E$ denotes the edges of it and $\mathcal{H}\in \mathbb{R}^{|V|\times|E|}$. Considering vertex $i$, if $i$ can be reached from edge $\epsilon$, it then can be denoted as $ \mathcal{H}_{i\epsilon} = 1$. Otherwise, $\mathcal{H}_{i\epsilon} = 0$. 

In a hypergraph, each hyperedge can encompasses more than two vertices, meaning:
\begin{gather}
\underset{n\in V}{\sum} \mathcal{H}_{n\epsilon} \in \mathbb{N}^{+}
\end{gather}
We use $W_\mathcal{H}$ to denote a diagonal matrix, the element of which represents the weight of each hyperedge. $D_v$ is the diagonal matrix which denotes the degree of each vertex and $D_e$ is the diagonal matrix which denotes the degree of each hyperedge as follows:

\begin{gather}
    D_{v,ii}=\sum_{\epsilon=1}^{|E|} W_{\mathcal{H},\epsilon\epsilon} \mathcal{H}_{i \epsilon} \\
    D_{e,\epsilon\epsilon}=\sum_{n=1}^{|V|} \mathcal{H}_{n \epsilon}
\end{gather}


Next, we introduce the algorithm to generate hyperedges for the constructed invocation graph, so as to learn the multivariate causality information for root cause analysis with hypergraph convolution. The trace typology graph illustrated in Fig.~\ref{pipeline}(c) is generated by removing all the log nodes and metric nodes from the multimodal invocation graph as the anomaly embedding are learnt at this stage from Equation~\ref{anomaly_emb}. Suppose an extra intervention $T$, for instance, a 5-second network package loss of the server, is applied to a random instance $j$. Taking the output of instance $j$ as a random variable $Y_j$, then 
\begin{gather}
\begin{split}
P(Y_j) & = \underset{i\in \textbf{Descendant}(j)}{\forall}P(Y_j \mid do(T_i)) \\
P(Y_j) & \neq \underset{i\in \textbf{Ascendant}(j)}{\forall}P(Y_j \mid do(T_i))
\end{split}
\end{gather}

Here we take instance node with index 6 as an example to consider the effect (centered with red circle in Fig~\ref{pipeline}(c)). Denoting the output of instance 6 with random variable $Y_6$. Since the instances with index larger than 6, which are descendant nodes $i\in [7,11]$, are called by instance 6 and their states are considered to be resulted from $Y_6$, intervention on these instances will not change the cause instance output $Y_6$. However, $Y_6$ is not independent with the intervention on its ascendant nodes including intervention directly imposed on itself, namely $i\in [1, 6]$. This leads to the hyperedge construction process (See Algorithm~\ref{algo1}).

\begin{algorithm} 
\caption{Trace Level Causality Hyperedge Construction}
\label{algo1}
\begin{flushleft}
\textbf{INPUT:} The invocation graph $\mathcal{G}(\mathcal{I}, E)$\\
\textbf{OUTPUT:} Incidence matrix $\mathcal{H}$
\end{flushleft}

 \begin{algorithmic}[1]
 \State $\mathcal{H} \gets \textbf{0}$ \Comment{Incidence matrix initialized to zero matrix}
 \For {$v \in \mathcal{I}$}
 \For {$v_p\in \textbf{Parent}(v)$} \Comment{For each invocation trace that reaches the specific instance, create a hyperedge connecting all upstream instances on the invocation path}
 \State $h \gets \mathbf{0}^{1\times |\mathcal{I}|}$
 \State $h[v] \gets 1, h[v_p] \gets 1$
 \For {$v_a\in \textbf{Ascendant}(v_p)$}
 \State $h[v_a] \gets 1$
 \EndFor
 \EndFor
 \State $\mathcal{H} \gets \mathcal{H} \mathbin\Vert h$
 \EndFor
 \State $h \gets \mathbf{0}^{1\times |\mathcal{I}|}$
 \State $h[v] \gets 1$
 \For {$v_d \in \textbf{Descendant}(v)$}
 \Comment{Create the hyperedge connecting all downstream instances}
 \State $h[v_d] \gets 1$
 \EndFor
 \State $\mathcal{H} \gets \mathcal{H} \mathbin\Vert h$
 \State \textbf{return} $\mathcal{H}$\Comment{The incidence matrix of the constructed hypergraph is $\mathcal{H}$}
 \end{algorithmic} 
\end{algorithm}

To be more specific, Algorithm~\ref{algo1} initializes the incidence matrix to an empty zero matrix with line 1. Line 2 loops through all the vertices of the trace invocation graph $\mathcal{G}$, picking a single vertex each time and constructing multiple hyperedges representing both causality and results. Lines 3-10 loop through each parent node of the target instance node, and construct a hyperedge that connects all ascendants of the parent node. Lines 12-17 construct the hyperedge that connects all the descendant nodes. Fig.~\ref{pipeline}(c) gives an illustration of the hyperedge construction for instance with index 6, which is marked in red. As instance 6 has two parent nodes, namely instances 2 and 5, a total number of three hyperedges will be generated for instance 6. 

With the hypergraph that represents the invocation causality being constructed, we initialize the node embeddings with $\Tilde{X}$ learnt from Equation~\ref{anomaly_emb}. In order to localize the trace level root cause instance, a convolution operation is performed on the hypergraph so that the causality information can propagate to each instance node from the constructed hyperedges. We consider each hyperedge as of the same significance, hence $W_\mathcal{H} = I$. And the hypergraph convolution can be defined as follows:
\begin{gather}
    \Tilde{X}^{(l)} = \sigma(D_v^{-\frac{1}{2}}\mathcal{H}D_e^{-1}\mathcal{H}^TD_v^{-\frac{1}{2}}\Tilde{X}^{(l-1)}\Theta^{l})
\end{gather}
We set $\Tilde{X^{(0)}} = \Tilde{X}$ from Equation~\ref{anomaly_emb}, and $\Theta \in \mathbb R^{d\times d^{\prime}}$ is the learnable projection that captures the causality of the trace level anomaly. In all, the root cause can be localized with a detection head matrix $D \in \mathbb R^{d^{\prime}\times 1}$ being applied to every instance node of the trace, and an end-to-end training schema can be accomplished by minimizing the overall loss:

\begin{gather}
\mathbb{L} = -\frac{1}{N}\sum_{i=1}^{N}(\sum_{n=1}^{|\mathcal{I}|}y_n^{(i)}\log(\text{Softmax}(\Tilde{X}^{(i,l)}D)_n))
\end{gather}

%% file: experiment.tex
\section{Experiments}\label{section4}
We evaluate the performance of CHASE framework in this section.  
Specifically, we compare the proposed framework with several baseline frameworks for RCA, including non-GNN based ones and GNN-based ones.  
To provide a comprehensive evaluation, we conduct experiments on two open-source datasets.    

\subsection{Experimental Setting}


\subsubsection{DataSets} 
The first dataset is a public dataset named Generic AIOps Atlas, provided by Cloudwise\footnote{https://github.com/CloudWise-OpenSource/GAIA-DataSet}; the second dataset is a real-world dataset collected from the AIOps 2020 competition\footnote{https://github.com/NetManAIOps/AIOps-Challenge-2020-Data}. We give the details of the datasets here. 

a) GAIA - The Generic AIOps Atlas dataset contains multi-modal information from the business simulation system MicroSS, including metrics, logs and traces. 
This dataset contains mobile service, log service, web service, database service and Redis service.  
It contains 10 service instances in total.  
The anomalies are injected into the log to simulate the malfunction of service invocation. 
The dataset contains four types of anomalies, including login failure, memory anomalies, access denied exceptions and missing files. This dataset includes 1099 static traces, which means the number of instance involved in each trace, the invocation order and trace typology all remain the same. Following the training and testing set split setting from DiagFusion~\cite{zhang2023robust}, we assign 160 traces for training, and the remaining for validation and testing.

b) AIOps 2020 - This dataset is collected for the AIOps 2020 challenge hosted by Tsinghua Netman Lab.  
The challenge aims at testing the availability of all microservices before releasing them into the production environment. This environment encompasses hundreds of microservice instances, including network instances, kernel instances, docker instances, etc. 
Each microservice instance is deployed on multiple physical machines, resulting in a highly complex system environment. Moreover, since each request can be handled by a subset of components of the overall system without involving all instances, invocation traces are considered dynamic. In other words, the trace typology in the dataset differ from one another. 
Fig.~\ref{1-1} is a service invocation sample from the dataset, with both log and metric data (CPU and memory usage, etc.) recorded for each service instance. Message instances trigger asynchronous invocations, represented with red arrows in the graph, while network gateway instances, service instances and database instances trigger synchronous invocations, represented with black arrows.  
For data collection, the whole system had been running for 3 months, in which 68 manually injected failures occurred, and each of them lasted for 5 minutes approximately. During each failure, there are hundreds of traces continuously being deployed in the system, among which may or may not be erroneous since a single failure would not result in the collapse of the whole microservice system. In order to evidently illustrate the performance of both baseline models and the proposed CHASE, we split each anomaly duration into a five-minute span, a three-minute span and a one-minute span based on the true label which records the starting timestamp of each failure, and we infer that a better approach can detect a higher percentage of anomaly traces within these time spans. As each failure is gradually recovered after its injection, the five-minute span potentially contains a lower percentage of anomaly traces compared with the one-minute span since the failure has a higher probability of being recovered and resulting in more non-anomaly traces.


\subsubsection{Baseline Methods}
We compare our proposed framework with seven baseline methods described as follows. The compared baselines can be divided into three categories: causality-based RCA approaches (i.e., PC and GES), correlation learning based approaches (i.e., CloudRanger and MicroRCA), and GNN based approaches (i.e., CausalRCA, TrinityRCL and Diagfusion). 

\begin{itemize}
\item \textbf{PC}~\cite{sole2017survey} is widely used for causal relationship inference and proven effective in identifying the root causes of system failures, process deviations, sensor failures, and insurance claims. It has been adopted in root cause analysis across various domains, such as system failure diagnosis, semiconductor manufacturing, wind turbines, and insurance.
\item \textbf{GES}~\cite{maxwell2002optimal} stands for Greedy Equivalence Search algorithm, which is mainly applied for casual relationship inference. Different from PC which relies on test of independence, GES applies optimal structure identification with greedy search to generate the causal graph.
\item \textbf{CloudRanger} is a root cause identification tool first introduced in~\cite{wang2018cloudranger}. It collects and analyzes system logs, metrics, and other data sources to build a machine learning model for root cause prediction.
\item \textbf{MicroRCA}~\cite{wu2020microrca} is an automated, fine-grained root cause localization framework to analyze monitoring data and localizes faulty services in the microservice architecture. It designs a gradient-based causal structure learning to capture linear and non-linear causal relations in monitoring metrics.
\item \textbf{TrinityRCL}~\cite{gu2023trinityrcl} localizes the root causes of anomalies at multiple levels of granularity by harnessing all types of telemetry data to construct a causal graph representing the intricate, dynamic, and nondeterministic relationships among the various entities related to the anomalies.
\item \textbf{CausalRCA}~\cite{xin2023causalrca} uses causal inference to automatically identify the root cause of performance issues in microservices. Specifically, it constructs a directed acyclic graph (DAG) that represents the causal relationships between services and their performance metrics, from which the root cause can be identified by inferring the causal relationships with the trained graph neural network. 
\item \textbf{Diagfusion}~\cite{zhang2023robust} is a robust failure diagnosis method that leverages multimodal data. It combines deployment data and traces to build a dependency graph, applies GNN to generate learnt representation for each service instance and achieve two-fold failure diagnosis, i.e., root cause instance localization and failure type determination.
\end{itemize}

Note that we apply the same hyperparameters of all the baseline RCA frameworks according to the best performance reported in their papers and source code. 
Both datasets are split into training/validation/testing sets, avoiding the possible overfitting issue. For baseline methods that require a PageRank to infer the root cause instance, we employ the damping factor as 0.85, maximum iteration as 100 and maximum error tolerance as 0.01. For CHASE, we apply 3 attention layers for heterogeneous message passing, each layer with 8 attention heads. We set LeakyReLU as the activation function with a negative slope coefficient being 0.3. We apply a single hypergraph convolution layer for causality learning, with embedding dimensionality equal to 128. The weight hyperparameter $\gamma$ in Equation.~\ref{anomaly_emb} is set to 0.5 by default.

\begin{table*}[t!]
\caption{Overall Performance Comparison}
\centering
\resizebox{\linewidth}{!}{%
\begin{tabular}{|c|c|c|c|c|c|c|c|}
\hline
\multicolumn{2}{|c|}{Dataset} & \multicolumn{3}{c}{GAIA} & \multicolumn{3}{|c|}{AIOps 2020} \\ \hline
\multicolumn{2}{|c|}{Evaluation Metric} & A@1 & A@3 & Avg@5 & Percentage@5 & Percentage@3 & Percentage@1 \\ \hline
\multirow{2}{*}{Causality-based} 
& PC & 0.2960 & 0.6368 & 0.5953 & - & - & - \\ \cline{2-8} 
& GSE & 0.3003 & 0.5399 & 0.5154 & 0.11 & 0.08 & 0.14 \\ \hline
\multirow{3}{*}{Correlation-based} 
& CloudRanger & 0.3290 & 0.4771 & 0.4883 & - & - & - \\ \cline{2-8} 
& MicroRCA & 0.3421 & 0.5528 & 0.5712 & 0.08 & 0.13 & 0.17 \\ \cline{2-8} 
& TrinityRCL & 0.4503 & 0.8244 & 0.7651 & 0.12 & 0.14 & 0.15 \\ \hline
\multirow{3}{*}{GNN-based} 
& CausalRCA & 0.3652 & 0.4973 & 0.5966 & - & - & - \\ \cline{2-8} 
& DiagFusion & 0.4121 & 0.8157 & 0.7484 & 0.12 & 0.14 & 0.14 \\ \cline{2-8} 
& \textbf{Proposed Method} & \textbf{0.6135} & \textbf{0.8823} & \textbf{0.8276} & \textbf{0.15} & \textbf{0.16} & \textbf{0.22} \\ \hline
\end{tabular}
}
\label{table:ER}
\begin{tablenotes}
\item {\scriptsize $-$} indicates baseline frameworks being inapplicable to certain dataset due to dynamic trace typology.
\end{tablenotes}
\end{table*}

\subsubsection{Evaluation Metrics}
Our framework aims to infer the root cause instance of each anomaly case accurately. 
We adopt the widely used evaluation metrics in the root cause inference task~\cite{zhang2023robust} while slightly modifying these metrics to suit real-world scenarios. We choose Top-1 accuracy ($A@1$), Top-3 accuracy ($A@3$) and Top-5 average accuracy ($Avg@5$) as three different evaluation metrics. 
Since often only the top three predicted results would be manually examined in real-world cases, we thus include $A@1$, $A@3$ and $Avg@5$ to evaluate the robustness of the root cause analysis methods.

Top-k accuracy ($A@k$) is calculated in the following way, with $RC_{a}$ as the ground truth root cause instance of anomaly trace $a$, $RC_{\hat{a}} [k]$ as the Top-k root cause instances set generated by the root cause inference system based on the information of anomaly trace $a$, $|D|$ as the size of the dataset:

\begin{equation}
A @ k=\frac{1}{|D|} \sum_{a \in D} \begin{cases}1, & \text { if } R C_{a} \in R C_{\hat{a}}[k] \\ 0, & \text { otherwise }\end{cases}
\end{equation}

$Avg@5$ is calculated based on the Top-k accuracy with k ranging from 1 to 5:

\begin{equation}
A v g @ 5=\frac{1}{5} \sum_{1 \leq k \leq 5} A @ k
\end{equation}

Percentage of erroneous traces occurred in $n$-minute span (Percentage@$n$) is calculated with the number of traces with being detected as anomalous divided by the total number of traces occurred within $n$ minutes after the labelled timestamp $T$, where $f(\cdot)$ denotes the binary prediction result:
\begin{equation}
\text{Percentage@}n=\frac{\underset{{x_t, t\in [T, T+n]}}{\sum}\mathds{1}\{f(x_t) = 1\}}{|\{x_t, t\in [T, T+n]\}|}
\end{equation}

\subsection{Evaluation Results}

To demonstrate the effectiveness of our proposed method, we compare CHASE with the baseline frameworks from different categories. The evaluation results are shown in Table~\ref{table:ER}.

The causality-based methods extract the causal graph from history traces, and both PC and GES apply PageRank on the causal graph with all edge weights equal to the out-degree of nodes. It can be observed clearly that the performance of PC and GES is worse than correlations-based methods, which generate the edge weights in the causal graph by calculating the correlations through monitored metrics. Specifically, MicroRCA integrates service invocation graphs with causal graphs and applies customized random walks, which outperforms CloudRanger that only considers the attributed graph with  anomalous propagation edges, indicating the necessity of leveraging the causal relationships. Regarding GNN-based methods, CausalRCA calculates edge weights through DAG-GNN with variational inference, requiring no training data labels. Due to the limitations of unsupervised learning, it reports worse performance to DiagFusion across all evaluation metrics as DiagFusion takes advantage of multi-modal data to generate the node features in the graph and trains end-to-end supervised RCA tasks. 
In general, our proposed method, which integrates microservice invocation graphs, anomaly heterogeneous graph and adaptively learnt causality with hypergraph, can obtain better performance compared with all baselines. On GAIA dataset, our proposed method can achieve 36.2\%, 7.2\%, 8.1\% higher than the best baseline on A@1, A@3, Avg@5, respectively. 

When it comes to the more complex AIOps 2020 dataset, the invocation typology of traces in training and testing set varies. As a result, the PC, CouldRanger and CausalRCA, which require static trace typology and are not inductive on traces with unseen typology from the training set, are not applicable to perform root cause analysis under such condition. Therefore, evaluation results of the aforementioned methods on AIOps 2020 dataset are left blank in Table~\ref{table:ER}. Our proposed method can still retain a significant performance gain of 25.0\%, 14.3\%, 29.4\% compared with the best baseline on the percentage of erroneous traces detected in 5-min span, 3-min span and 1-min span.


\begin{figure*}[t]
\includegraphics[width=1\textwidth]{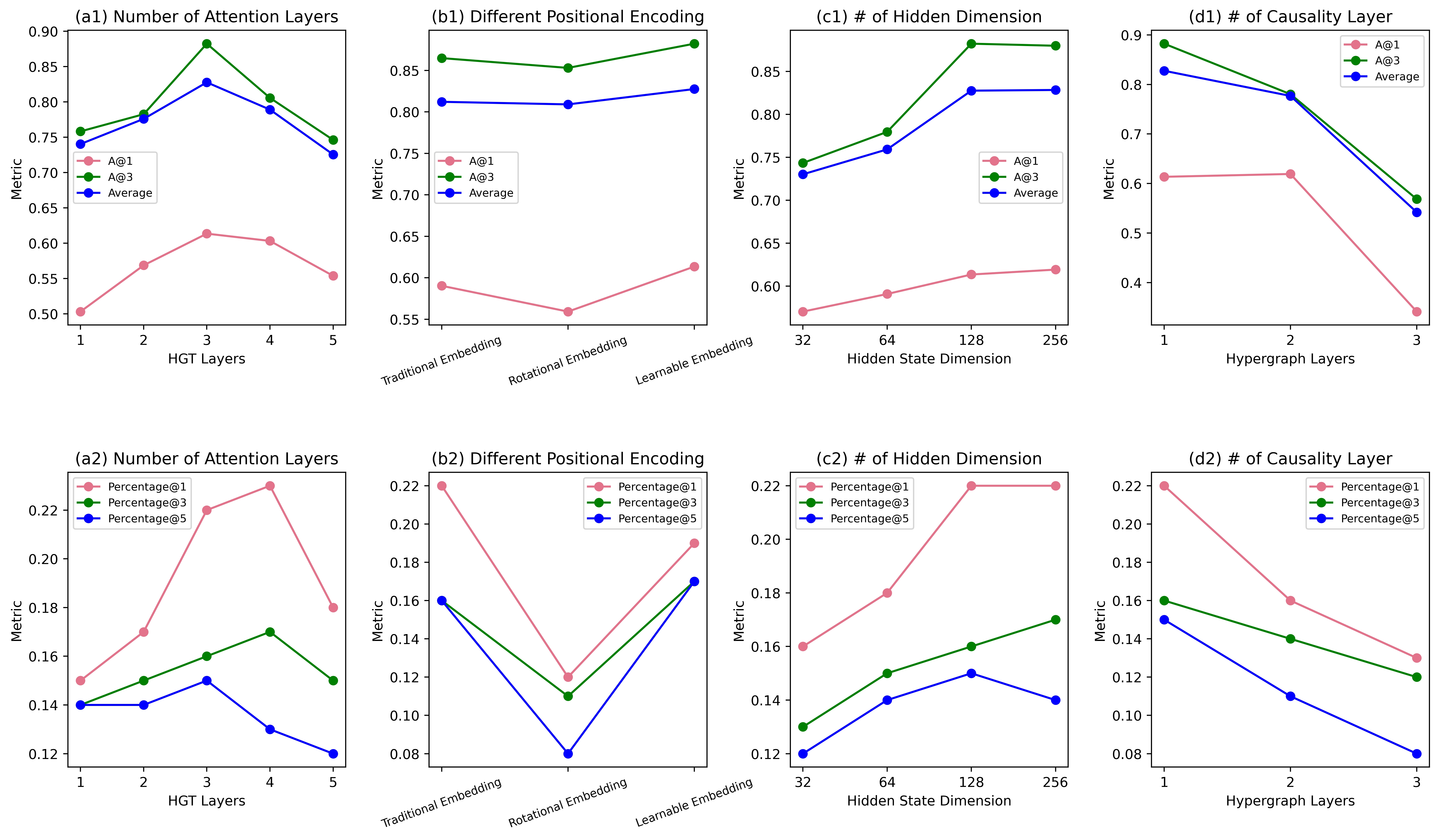}
\centering
\caption{Sensitivity analysis: (a) number of attention layers; (b) different positional encoding; (c) hidden dimension; (d) number of causality layers} 
\label{4.2-5}
\end{figure*}

\subsection{Sensitivity Analysis}

We conduct the sensitivity analysis in this part on both datasets. Specifically, Fig.~\ref{4.2-5} (a1) and (a2) shows how the number of attention layers affects the performance on GAIA and AIOps 2020 dataset, respectively. When the number of attention layers increases up to 3, there is typically an improvement in the performance of our model. The performance starts to drop afterwards, indicating the potential overfitting problem. 
Regarding different types of positional encoding, depicted in Fig.~\ref{4.2-5} (b1) and (b2), traditional embedding and learnable embedding demonstrate superior performances. We apply tradition embedding as it is simply composed of constants calculated from Equation.\ref{eq:positional}, which reduce the total amount of learnable parameters and result in lower computational cost.
The effect of hidden state dimension for heterogeneous message passing and hypergraph convolution is depicted in Fig.~\ref{4.2-5} (c1) and (c2), the performance of the model evidently surges until the dimension reaches 128, then the performance is only slightly improved when hidden state dimension is doubled to 256. Thus, considering the trade-off between computational efficiency and performance, we apply a hidden state dimension of 128.
For the number of hypergraph convolution layers demonstrated in Fig.~\ref{4.2-5} (d1) and (d2), the performance of the model continues to drop as the number of layer increases. Hence, we apply a single hypergraph convolution layer. Intuitively, as a single hypergraph layer is already capable of capturing high-order multivariate correlation, stacking up hypergraph layers brings in extra redundancies in the training process, thus causing undesirable performance drops.

\subsection{Ablation Study}
\begin{table}[t]
\caption{Ablation study results}
\centering
\begin{tabular}{|c|c|c|c|}
\hline
Ablation Strategy                                     & A@1    & A@3    & Avg@5  \\ \hline
Default Method                                      & 0.6135 & 0.8823 & 0.8276 \\ \hline
V1: w/o instance embedding  & 0.5927 & 0.8554 & 0.7852 \\ \hline
V2: w/o heterogeneous message passing & 0.3845 & 0.5403 & 0.6581 \\ \hline
V3: w/o causal hyperedge            & 0.4679 & 0.8106 & 0.7598 \\ \hline
\end{tabular}%
\label{table:AR}
\end{table}

To evaluate the effectiveness of the main components of CHASE, we conducted an ablation study on the GAIA dataset using the following three variations: 
V1: remove the instance embedding layer while keeping the non-learnable positional embedding layer. 
V2: remove the heterogeneous message passing from the graph convolution layer. 
V3: remove the final hypergraph layer which learns the propagation of causality.
Table \ref{table:AR} lists the change of performance of each variant. 

Since the instance embedding layer captures the difference between instances, reducing these features in V1 will result in missing information of the whole trace typology and invocation structure, leading to a subtle drop of performance of 3.3\% in A$@1$, clearly shown in Table \ref{table:AR}.

\begin{figure}[t]
\includegraphics[width=1\linewidth]{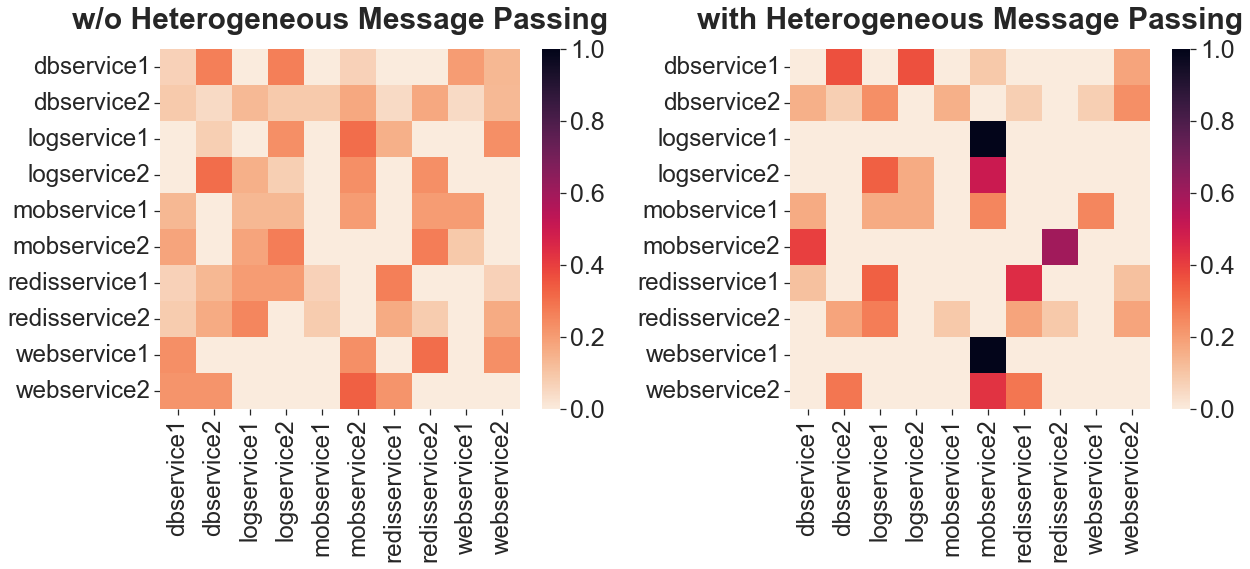}
\centering
\caption{Causal weights visualization}
\label{4.2-1}
\end{figure}

Considering the V2 variant of CHASE, we replace the heterogeneous message passing layer with a homogeneous graph attention layer, which results in a 37\% performance drop in A$@1$. This emphasizes the significance to model the heterogeneous correlation of the invocation graph. To provide a clearer illustration of the impact of learning the heterogeneity attributes among metrics, logs, and instances through message passing in CHASE, Fig.~\ref{4.2-1} plots a comparison between the homogeneous edge weight and the heterogeneous edge weight on the $160^{th}$ trace in GAIA dataset (webservice2 as the root cause instance). As the edges related to webservice2 (last columns in both figures) are granted with higher weights, it indicates that the heterogeneous message passing layer is able to enhance the root cause inference's performance, which is achieved by directing more attention towards the edges related to the instance of the root cause.

In the context of causality inference, it has been observed that directly integrating the detection head into the output of the heterogeneous message passing layer in the V3 variant leads to the causal hypergraph degrading into a heterogeneous graph neural network. In this scenario, information propagates only through K hops (where K represents the number of layers), which contradicts the nature of anomaly causality that may propagate across an arbitrary number of hops depending on the invocation. This leads to a performance drop of 23.7\% in A$@$1. 


In comparison to the performance of V2, V3 which retains the heterogeneity is considerably more critical. As a result, V3 exhibits higher accuracy of 21.6\% in A$@$1 than V2, since causal edges provide less topological information than actual invocation edges. Nevertheless, V3 demonstrates superior robustness (33.3\% higher in A$@$3 and 13.3\% higher in Avg$@$5) than V2, as causal edges denote stronger causality than actual invocation edges.

%% file: main.bbl
\begin{thebibliography}{10}
\providecommand{\url}[1]{#1}
\csname url@samestyle\endcsname
\providecommand{\newblock}{\relax}
\providecommand{\bibinfo}[2]{#2}
\providecommand{\BIBentrySTDinterwordspacing}{\spaceskip=0pt\relax}
\providecommand{\BIBentryALTinterwordstretchfactor}{4}
\providecommand{\BIBentryALTinterwordspacing}{\spaceskip=\fontdimen2\font plus
\BIBentryALTinterwordstretchfactor\fontdimen3\font minus \fontdimen4\font\relax}
\providecommand{\BIBforeignlanguage}[2]{{%
\expandafter\ifx\csname l@#1\endcsname\relax
\typeout{** WARNING: IEEEtran.bst: No hyphenation pattern has been}%
\typeout{** loaded for the language `#1'. Using the pattern for}%
\typeout{** the default language instead.}%
\else
\language=\csname l@#1\endcsname
\fi
#2}}
\providecommand{\BIBdecl}{\relax}
\BIBdecl

\bibitem{li2021practical}
Z.~Li, J.~Chen, R.~Jiao, N.~Zhao, Z.~Wang, S.~Zhang, Y.~Wu, L.~Jiang, L.~Yan, Z.~Wang, Z.~Chen, W.~Zhang, X.~Nie, K.~Sui, and D.~Pei, ``Practical root cause localization for microservice systems via trace analysis,'' in \emph{International Symposium on Quality of Service}.\hskip 1em plus 0.5em minus 0.4em\relax IEEE, 2021, pp. 1--10.

\bibitem{baboi2019dynamic}
M.~Baboi, A.~Iftene, and D.~G{\^\i}fu, ``Dynamic microservices to create scalable and fault tolerance architecture,'' \emph{Procedia Computer Science}, vol. 159, pp. 1035--1044, 2019.

\bibitem{ramu2023performance}
V.~Ramu, ``Performance impact of microservices architecture,'' \emph{The Review of Contemporary Scientific and Academic Studies}, vol.~3, 2023.

\bibitem{zhang2019esda}
T.~Zhang, Z.~Shen, J.~Jin, A.~Tagami, X.~Zheng, and Y.~Yang, ``{ESDA:} an energy-saving data analytics fog service platform,'' in \emph{International Conference on Service-Oriented Computing}.\hskip 1em plus 0.5em minus 0.4em\relax Springer, 2019, pp. 171--185.

\bibitem{peng2022large}
X.~Peng, ``Large-scale trace analysis for microservice anomaly detection and root cause localization,'' in \emph{Proceedings of the Federated Africa and Middle East Conference on Software Engineering}, 2022, pp. 93--94.

\bibitem{liu2021microhecl}
D.~Liu, C.~He, X.~Peng, F.~Lin, C.~Zhang, S.~Gong, Z.~Li, J.~Ou, and Z.~Wu, ``Microhecl: High-efficient root cause localization in large-scale microservice systems,'' in \emph{2021 IEEE/ACM 43rd International Conference on Software Engineering: Software Engineering in Practice (ICSE-SEIP)}.\hskip 1em plus 0.5em minus 0.4em\relax IEEE, 2021, pp. 338--347.

\bibitem{chen2014causeinfer}
P.~Chen, Y.~Qi, P.~Zheng, and D.~Hou, ``Causeinfer: Automatic and distributed performance diagnosis with hierarchical causality graph in large distributed systems,'' in \emph{IEEE INFOCOM 2014-IEEE Conference on Computer Communications}.\hskip 1em plus 0.5em minus 0.4em\relax IEEE, 2014, pp. 1887--1895.

\bibitem{ikram2022root}
A.~Ikram, S.~Chakraborty, S.~Mitra, S.~Saini, S.~Bagchi, and M.~Kocaoglu, ``Root cause analysis of failures in microservices through causal discovery,'' \emph{Advances in Neural Information Processing Systems}, vol.~35, pp. 31\,158--31\,170, 2022.

\bibitem{brandon2020graph}
{\'A}.~Brand{\'o}n, M.~Sol{\'e}, A.~Hu{\'e}lamo, D.~Solans, M.~S. P{\'e}rez, and V.~Munt{\'e}s-Mulero, ``Graph-based root cause analysis for service-oriented and microservice architectures,'' \emph{Journal of Systems and Software}, vol. 159, p. 110432, 2020.

\bibitem{kim2013root}
M.~Kim, R.~Sumbaly, and S.~Shah, ``Root cause detection in a service-oriented architecture,'' \emph{ACM SIGMETRICS Performance Evaluation Review}, vol.~41, pp. 93--104, 2013.

\bibitem{soldani2022anomaly}
J.~Soldani and A.~Brogi, ``Anomaly detection and failure root cause analysis in (micro) service-based cloud applications: A survey,'' \emph{ACM Computing Surveys}, vol.~55, no.~3, pp. 1--39, 2022.

\bibitem{sole2017survey}
M.~Sol{\'e}, V.~Munt{\'e}s-Mulero, A.~I. Rana, and G.~Estrada, ``Survey on models and techniques for root-cause analysis,'' \emph{arXiv:1701.08546}, 2017.

\bibitem{maxwell2002optimal}
D.~M. Chickering, ``Optimal structure identification with greedy search,'' \emph{Journal of Machine Learning Research}, vol.~3, pp. 507--554, 2002.

\bibitem{lin2018microscope}
J.~Lin, P.~Chen, and Z.~Zheng, ``Microscope: Pinpoint performance issues with causal graphs in micro-service environments,'' in \emph{International Conference on Service-Oriented Computing}.\hskip 1em plus 0.5em minus 0.4em\relax Springer, 2018, pp. 3--20.

\bibitem{sharma2013cloudpd}
B.~Sharma, P.~Jayachandran, A.~Verma, and C.~R. Das, ``Cloud{PD}: Problem determination and diagnosis in shared dynamic clouds,'' in \emph{International Conference on Dependable Systems and Networks}.\hskip 1em plus 0.5em minus 0.4em\relax IEEE, 2013, pp. 1--12.

\bibitem{wang2018cloudranger}
P.~Wang, J.~Xu, M.~Ma, W.~Lin, D.~Pan, Y.~Wang, and P.~Chen, ``Cloud{R}anger: Root cause identification for cloud native systems,'' in \emph{International Symposium on Cluster, Cloud and Grid Computing}.\hskip 1em plus 0.5em minus 0.4em\relax IEEE, 2018, pp. 492--502.

\bibitem{wu2020microrca}
L.~Wu, J.~Tordsson, E.~Elmroth, and O.~Kao, ``Micro{RCA}: Root cause localization of performance issues in microservices,'' in \emph{Network Operations and Management Symposium}.\hskip 1em plus 0.5em minus 0.4em\relax IEEE, 2020, pp. 1--9.

\bibitem{xin2023causalrca}
R.~Xin, P.~Chen, and Z.~Zhao, ``Causal{RCA}: Causal inference based precise fine-grained root cause localization for microservice applications,'' \emph{Journal of Systems and Software}, vol. 203, 2023.

\bibitem{zhang2023robust}
S.~Zhang, P.~Jin, Z.~Lin, Y.~Sun, B.~Zhang, S.~Xia, Z.~Li, Z.~Zhong, M.~Ma, W.~Jin, D.~Zhang, Z.~Zhu, and D.~Pei, ``Robust failure diagnosis of microservice system through multimodal data,'' \emph{arXiv:2302.10512}, 2023.

\bibitem{rong2022locating}
G.~Rong, H.~Wang, S.~Gu, Y.~Xu, J.~Sun, D.~Shao, and H.~Zhang, ``Locating anomaly clues for atypical anomalous services: An industrial exploration,'' \emph{IEEE Transactions on Dependable and Secure Computing}, pp. 2746--2761, 2023.

\bibitem{bhandari2018fault}
G.~P. Bhandari and R.~Gupta, ``Fault analysis of service-oriented systems: A systematic literature review,'' \emph{IET Software}, vol.~12, no.~6, pp. 446--460, 2018.

\bibitem{zhang2024trace}
C.~Zhang, Z.~Dong, X.~Peng, B.~Zhang, and M.~Chen, ``Trace-based multi-dimensional root cause localization of performance issues in microservice systems,'' in \emph{Proceedings of the IEEE/ACM 46th International Conference on Software Engineering}, 2024, pp. 1--12.

\bibitem{yu2021microrank}
G.~Yu, P.~Chen, H.~Chen, Z.~Guan, Z.~Huang, L.~Jing, T.~Weng, X.~Sun, and X.~Li, ``Microrank: End-to-end latency issue localization with extended spectrum analysis in microservice environments,'' in \emph{Proceedings of the Web Conference 2021}, 2021, pp. 3087--3098.

\bibitem{gholami2021comparative}
A.~Gholami and A.~K. Srivastava, ``Comparative analysis of {ML} techniques for data-driven anomaly detection, classification and localization in distribution system,'' in \emph{North American Power Symposium}.\hskip 1em plus 0.5em minus 0.4em\relax IEEE, 2021, pp. 1--6.

\bibitem{zhang2021cloudrca}
Y.~Zhang, Z.~Guan, H.~Qian, L.~Xu, H.~Liu, Q.~Wen, L.~Sun, J.~Jiang, L.~Fan, and M.~Ke, ``Cloud{RCA}: a root cause analysis framework for cloud computing platforms,'' in \emph{Proceedings of the ACM International Conference on Information \& Knowledge Management}, 2021, pp. 4373--4382.

\bibitem{ma2020self}
M.~Ma, W.~Lin, D.~Pan, and P.~Wang, ``Self-adaptive root cause diagnosis for large-scale microservice architecture,'' \emph{IEEE Transactions on Services Computing}, vol.~15, pp. 1399--1410, 2020.

\bibitem{jia2021logflash}
T.~Jia, Y.~Wu, C.~Hou, and Y.~Li, ``Log{F}lash: Real-time streaming anomaly detection and diagnosis from system logs for large-scale software systems,'' in \emph{International Symposium on Software Reliability Engineering}.\hskip 1em plus 0.5em minus 0.4em\relax IEEE, 2021, pp. 80--90.

\bibitem{gu2023trinityrcl}
S.~Gu, G.~Rong, T.~Ren, H.~Zhang, H.~Shen, Y.~Yu, X.~Li, J.~Ouyang, and C.~Chen, ``Trinity{RCL}: Multi-granular and code-level root cause localization using multiple types of telemetry data in microservice systems,'' \emph{IEEE Transactions on Software Engineering}, vol.~49, no.~5, pp. 3071--3088, 2023.

\bibitem{kipf2016semi}
T.~N. Kipf and M.~Welling, ``Semi-supervised classification with graph convolutional networks,'' \emph{arXiv preprint arXiv:1609.02907}, 2016.

\bibitem{hamilton2017inductive}
W.~Hamilton, Z.~Ying, and J.~Leskovec, ``Inductive representation learning on large graphs,'' \emph{Advances in Neural Information Processing Systems}, vol.~30, 2017.

\bibitem{velickovic2017graph}
P.~Velickovic, G.~Cucurull, A.~Casanova, A.~Romero, P.~Lio, and Y.~Bengio, ``Graph attention networks,'' \emph{Statistics}, vol. 1050, no.~20, pp. 10--48\,550, 2017.

\bibitem{zhang2023adaptive}
T.~Zhang, Y.~Liu, Z.~Shen, R.~Xu, X.~Chen, X.~Huang, and X.~Zheng, ``An adaptive federated relevance framework for spatial temporal graph learning,'' \emph{IEEE Transactions on Artificial Intelligence}, vol.~5, pp. 2227--2240, 2024.

\bibitem{liu2024exploiting}
Y.~Liu, Z.~Zhao, T.~Zhang, K.~Wang, X.~Chen, X.~Huang, J.~Yin, and Z.~Shen, ``Exploiting spatial-temporal data for sleep stage classification via hypergraph learning,'' in \emph{IEEE International Conference on Acoustics, Speech and Signal Processing}.\hskip 1em plus 0.5em minus 0.4em\relax IEEE, 2024, pp. 5430--5434.

\bibitem{wang2023hierarchical}
D.~Wang, Z.~Chen, J.~Ni, L.~Tong, Z.~Wang, Y.~Fu, and H.~Chen, ``Hierarchical graph neural networks for causal discovery and root cause localization,'' \emph{arXiv preprint arXiv:2302.01987}, 2023.

\bibitem{lee2023eadro}
C.~Lee, T.~Yang, Z.~Chen, Y.~Su, and M.~R. Lyu, ``Eadro: An end-to-end troubleshooting framework for microservices on multi-source data,'' in \emph{2023 IEEE/ACM 45th International Conference on Software Engineering (ICSE)}.\hskip 1em plus 0.5em minus 0.4em\relax IEEE, 2023, pp. 1750--1762.

\bibitem{joulin2016bag}
A.~Joulin, E.~Grave, P.~Bojanowski, and T.~Mikolov, ``Bag of tricks for efficient text classification,'' \emph{arXiv:1607.01759}, 2016.

\end{thebibliography}
